\documentclass{article}

\usepackage[preprint]{neurips_2026}

\usepackage{microtype}
\usepackage{booktabs}
\usepackage{graphicx}
\usepackage{amsmath,amssymb}
\usepackage{hyperref}
\hypersetup{hidelinks}
\usepackage{xspace}

\newcommand{\sys}{\textsc{LearnActCoder}\xspace}

\title{LearnActCoder: Role-Aware Error Memory for Adaptive Clinical Coding Agents}

\author{%
  Meysam Ghaffari$^{1}$, Bhaskar Sen$^{1}$, Nasim Sabetpour$^{1}$, Nina Fatehi$^{1}$,
  Animesh Agarwal$^{1}$, Carlos Morato$^{1,2}$ \\[4pt]
  $^{1}$Optum AI \qquad $^{2}$Harvard University
}

\begin{document}
\maketitle

\begin{abstract}
Clinical coding agents repeatedly encounter the same failure modes, including unsupported
codes, missed documented conditions, specificity errors, and procedure-coding convention
mismatches. We introduce \textbf{Learn-Then-Act}, an inference-time adaptation framework
that converts errors from a small labeled LEARN batch into a structured \emph{Mistake
Knowledge Database} (MistakeKDB). False-negative lessons are routed to a recall-oriented
Coder, while false-positive lessons are routed to a precision-oriented Judge. We
instantiate the framework in \sys{}, a Coder--Judge clinical coding pipeline with
lookup-table grounding where available. On 150 matched MIMIC-III notes, structured
MistakeKDB improves CPT F1 by 5.9 percentage points, while raw-example and
reflection-style memories remain near the no-memory baseline; the ICD-9 improvement is
not significant. On a matched MIMIC-IV cohort, memory shifts ICD-10 coding toward higher
precision at a recall cost, leaving F1 statistically unchanged. Applying the same memory
to 1{,}000 held-out MIMIC-III notes maintains a stable ICD operating point, providing
scale/stability evidence. Overall, the results are consistent with structured,
feedback-derived error memory being useful for adapting clinical coding behavior across
cases without weight updates or changes to the underlying workflow. Absolute CPT/HCPCS
performance remains low, and the system is evaluated retrospectively rather than in
clinical deployment.
\end{abstract}

\section{Introduction}
\label{sec:intro}

Medical coding maps clinical documentation to standardized diagnosis and procedure
vocabularies such as ICD and CPT/HCPCS. The resulting codes affect reimbursement,
utilization measurement, quality reporting, audit exposure, and downstream clinical
research~\citep{OMalley2005}. The task is difficult for language models because a note
contains many clinically plausible concepts that are not necessarily codable, code
specificity depends on fine-grained documentation, and the output space contains tens of
thousands of labels.

Recent LLM systems approach coding through prompting, retrieval, staged reasoning, or
multi-agent workflows~\citep{Boyle2023GPTCoding,yang2023surpassing,Huang2024LLMCoding,Huang2024CodeLikeHumans}.
These methods can improve a single prediction, but a separate operational problem
remains: \emph{how should an agent reuse corrected errors across subsequent cases without
updating model weights?} In practice, coding mistakes are often recurrent. A system may
repeatedly over-code weakly supported diagnoses, miss status/history codes, choose an
incorrect specificity level, or confuse professional and facility procedure-code
conventions. Treating each note as an independent zero-shot episode discards this
feedback.

We study a lightweight form of post-error adaptation. \textbf{Learn-Then-Act} converts
prediction/reference-label mismatches from a small LEARN batch into reusable natural-
language lessons. Its key design choice is \emph{role-aware routing}: lessons derived from
false negatives are supplied to the recall-oriented Coder, whereas lessons derived from
false positives are supplied to the precision-oriented Judge. The underlying
Coder--Judge workflow and model weights remain fixed. This distinguishes the method from
within-episode self-refinement and from approaches that learn or search over agentic
workflows.

We instantiate Learn-Then-Act in \sys{}, a clinical coding agent that proposes code sets,
checks candidate codes against a reference lookup database where coverage is available,
and revises predictions through Coder--Judge interaction. We evaluate the method on
MIMIC-III and MIMIC-IV with held-out and matched-note designs. The evidence supports a
narrower but practically relevant claim than uniform accuracy improvement: mistake
memory can systematically shift precision/recall behavior, and in a 150-note matched
MIMIC-III analysis it yields a significant relative improvement in CPT F1. On MIMIC-IV,
the same mechanism improves ICD precision at a measurable recall cost, making the
trade-off explicit rather than hiding it behind a single aggregate score.

\paragraph{Contributions.}
\begin{enumerate}
  \item \textbf{Role-aware error memory.} We introduce Learn-Then-Act, which distills
    recurring coding errors into structured lessons and routes false-negative and
    false-positive knowledge to agents with complementary recall and precision roles,
    without fine-tuning.
  \item \textbf{Grounded clinical-coding instantiation.} We implement the framework in
    \sys{}, combining a recall-oriented Coder, a precision-oriented Judge, and
    lookup-table validation that separates code existence/description checks from
    clinical support.
  \item \textbf{Matched, representation, and scaled evaluation.} Across MIMIC-III and
    MIMIC-IV, we report matched-note memory ablations, a four-way $n{=}150$
    memory-representation comparison on identical notes, paired uncertainty estimates,
    a 1{,}000-note stability evaluation, an end-to-end inference-latency analysis,
    component analyses, and an error taxonomy that exposes both memory-addressable errors
    and limitations caused by missing documentation or billing context.
\end{enumerate}

\section{Related Work}
\label{sec:related}

\paragraph{Experience and feedback for LLM agents.}
Reflexion~\citep{Shinn2023Reflexion} and Self-Refine~\citep{Madaan2023SelfRefine} use
language feedback to revise reasoning, while ExpeL~\citep{Zhao2024ExpeL} accumulates
experience across tasks. Learn-Then-Act is closest in spirit to cross-instance
experience reuse, but stores error-specific, typed lessons and routes them according to
agent role. The contribution is therefore not generic textual memory, but the mapping
between an observed error direction (false negative or false positive), a structured
coding lesson, and the agent responsible for recall or precision.

\paragraph{Automatic and LLM-based clinical coding.}
Supervised ICD coding is commonly formulated as extreme multi-label classification,
with systems including CAML~\citep{Mullenbach2018}, MultiResCNN~\citep{Li2020MultiResCNN},
LAAT~\citep{Vu2020LAAT}, PLM-ICD~\citep{Huang2022PLMICD}, ICD-MSMN~\citep{Yuan2022MSMN},
KEPT~\citep{Yang2023KEPT}, ICDXML~\citep{Wang2024ICDXML}, ACE-ICD~\citep{Ren2025ACEICD},
and recent label-space or retrieval methods~\citep{RoSimTail2025,gomes2024accurate,boukhers2024large}.
LLM-based coding offers a different trade-off: it can operate with little or no
 task-specific training and can incorporate natural-language guidelines, but it remains
susceptible to unsupported or malformed outputs~\citep{Ji2023Hallucination,Pal2023MedHALT}.
Open CPT/HCPCS evaluation is additionally complicated by licensing and by differences
between professional and facility coding conventions~\citep{Xerk2024MedCodingAI}.

\paragraph{Multi-agent and adaptive coding workflows.}
Role-specialized debate has been explored for factuality and reasoning~\citep{Du2023Debate,Liang2023Divergent},
and medical multi-agent systems motivate decomposing generation and review~\citep{Tang2024MedAgents}.
Coding-specific work is increasingly agentic. Code Like Humans (CLH) implements a staged,
guideline-oriented workflow and demonstrates inference over the full ICD-10 code
space~\citep{Huang2024CodeLikeHumans}. Most directly related, MedDCR~\citep{Zheng2026MedDCR}
treats the \emph{workflow} as the object of learning: a Designer proposes workflows, a
Coder executes them, a Reflector evaluates them, and a memory archive stores prior
workflow designs. Learn-Then-Act instead keeps the workflow fixed and adapts the
\emph{error knowledge supplied to each role}. This distinction lets us test whether a
small audit batch can change subsequent behavior without workflow search or weight
updates.

\section{Method}
\label{sec:method}

\subsection{Base Pipeline: Generate, Ground, and Review}

\sys{} uses two inference agents plus a non-generative lookup component and a post-hoc
MistakeAnalyzer. The Coder proposes a broad candidate set; CodeLookupDB checks candidate
identifiers and reference descriptions where coverage exists; the Judge reviews clinical
support and specificity; and the MistakeAnalyzer is used only after reference labels are
available in the LEARN phase. Separating these stages makes it possible to direct recall-
and precision-oriented feedback to different components.

\paragraph{Coder agent (recall-oriented).}
Given a discharge note, the Coder produces structured JSON containing candidate codes,
descriptions, confidence scores, and supporting evidence. Its prompt favors recall and
performs an initial extraction followed by a category-level ``what did I miss?'' pass.
The Coder is therefore the natural recipient of lessons derived from false negatives.

\paragraph{CodeLookupDB grounding.}
Before Judge review, candidate identifiers are checked against a lookup database. In the
base-pipeline benchmark configuration, the database contains 192{,}214 ICD-10-CM entries
from a UMLS-derived source and 26{,}533 CPT/HCPCS entries with descriptions and
hierarchical categories. The report marks a candidate as present or not found in the
loaded table and exposes the reference description to the Judge. This check addresses
nonexistent or description-mismatched codes; it does \emph{not} establish that a present
code is clinically warranted. Direct ICD-9 lookup is disabled because the loaded ICD
reference is ICD-10 based.

\paragraph{Judge agent (precision-oriented).}
The Judge receives the note, Coder proposals, and lookup report. It prioritizes lookup
status where available, note-level evidence, agreement between the code description and
clinical context, specificity, and output-format compliance. It can accept the proposal
or request a revised complete code set. Because the Judge is designed to remove weakly
supported candidates, it is the primary recipient of lessons derived from false
positives.

\paragraph{Debate protocol.}
Coder generation, lookup, and Judge review repeat for at most $R{=}3$ rounds. This
creates a generate--filter architecture: the Coder searches broadly, while the Judge
controls precision. Importantly, the same architecture is used with and without
MistakeKDB in the matched memory ablations.

\subsection{Learn-Then-Act: Error Memory Across Cases}
\label{sec:learn-then-act}

Learn-Then-Act has two phases. The LEARN phase converts errors on labeled cases into a
compact memory; the ACT phase reuses that memory on unseen notes without modifying model
weights.

\paragraph{Phase 1: LEARN.}
After the base pipeline codes each of $N_L$ LEARN notes, the MistakeAnalyzer receives the
note, predictions, and reference administrative codes. For every false positive and
false negative, it records an explanation and a transferable lesson and assigns one of
six operational categories: \emph{over-coding}, \emph{under-coding}, \emph{documentation
gap}, \emph{specificity error}, \emph{wrong code type}, or \emph{other}. The resulting
entries form MistakeKDB, a structured JSON repository indexed by error category, code
family, and clinical context. Because these labels are generated by the MistakeAnalyzer,
we treat the taxonomy as a mechanistic analysis rather than expert adjudication.

\paragraph{Phase 2: ACT and role-aware routing.}
For subsequent notes, false-negative lessons are injected into the Coder prompt to
expand category search, whereas false-positive lessons are injected into the Judge
prompt to tighten evidence requirements. For example, a recurring missed-code lesson
may instruct the Coder to inspect chronic/status categories, while an over-coding lesson
may instruct the Judge to reject diagnoses supported only by differential or passing
mentions. In shorthand,
\[
  \mathcal{M}_{\mathrm{FN}} \rightarrow \text{Coder}, \qquad
  \mathcal{M}_{\mathrm{FP}} \rightarrow \text{Judge}.
\]
The routing is deliberately asymmetric: it aligns the direction of the observed error
with the role responsible for correcting that error type.

\begin{figure}[t]
\centering
\includegraphics[width=\linewidth]{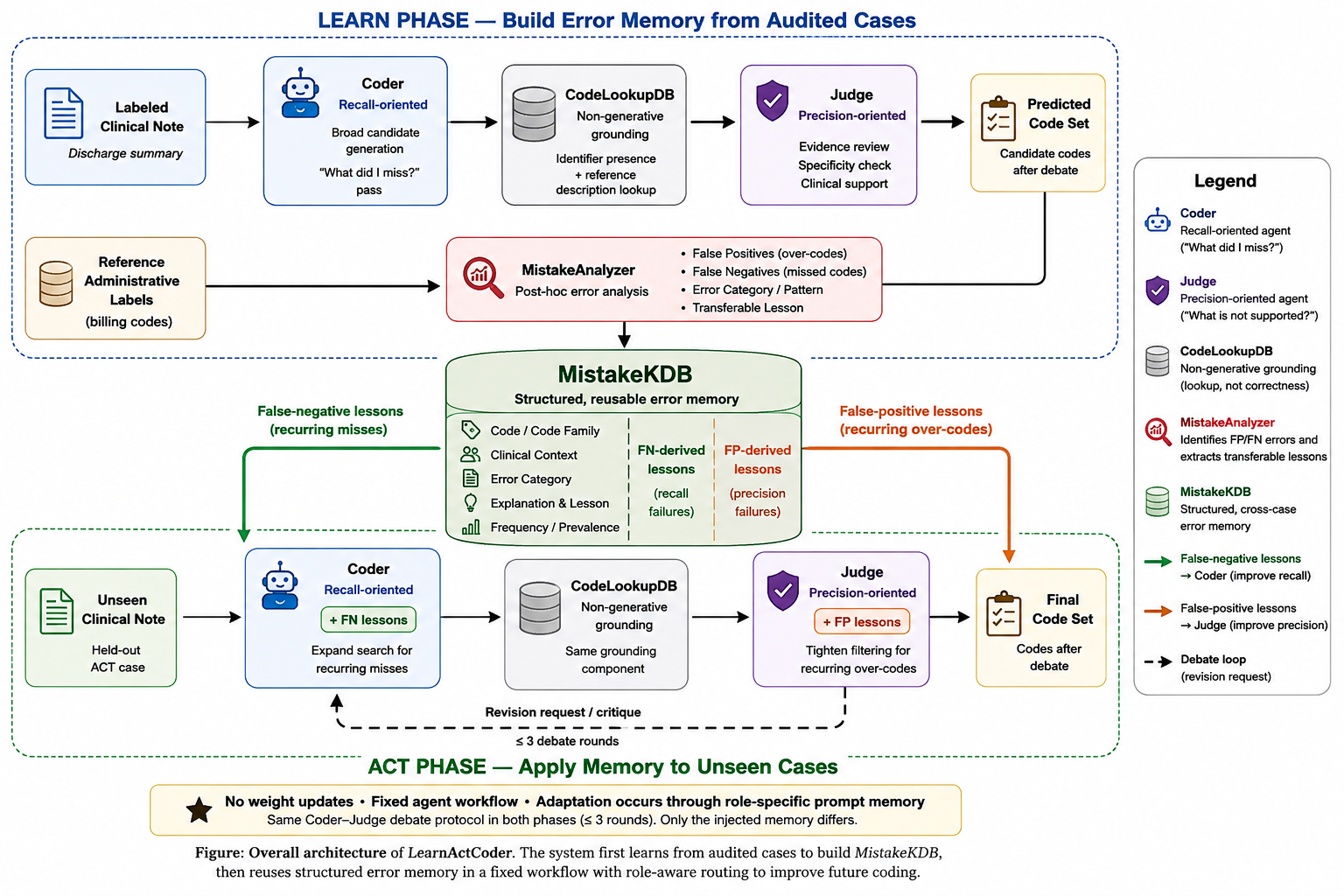}
\caption{\textbf{Learn-Then-Act architecture.} During LEARN, predictions and reference
administrative labels are analyzed into structured error memory. During ACT,
false-negative (FN) lessons are routed to the recall-oriented Coder and false-positive
(FP) lessons to the precision-oriented Judge. The lookup component, model weights, and
Coder--Judge workflow remain fixed; only role-specific prompt memory changes across
cases.}
\label{fig:architecture}
\end{figure}

Figure~\ref{fig:architecture} summarizes the complete cross-case feedback loop and
highlights the asymmetric routing that distinguishes Learn-Then-Act from ordinary
within-note refinement.

\paragraph{Memory representation.}
Each MistakeKDB entry stores the error category, implicated code(s), a natural-language
explanation, and a transferable lesson. Similar lessons are consolidated and frequent
patterns receive higher priority before prompt injection. The current implementation is
therefore a compact error memory rather than retrieval over full prior notes. In the
memory-format study, all conditions use the same underlying LEARN-phase error pool
(1{,}028 prediction/reference-label mismatches), the same 150 ACT notes, and the same
inference pipeline; they differ in how that error experience is represented. Thus, the
ablation compares alternative representations of the same underlying error pool rather
than identical prompt text.

\section{Experiments}
\label{sec:experiments}

\paragraph{Datasets and labels.}
We use MIMIC-III~\citep{Johnson2016MIMICIII} for ICD-9 + CPT experiments and
MIMIC-IV~\citep{Johnson2023MIMICIV} for ICD-10 + HCPCS experiments. Reference labels are
derived from the corresponding diagnosis, procedure, and CPT/HCPCS event tables. We call
them \emph{reference administrative labels} rather than clinical ground truth because
billing records can encode information not stated in an individual discharge summary.
The main MIMIC-III LEARN batch contains 25 labeled discharge summaries. ACT evaluation is
note-level held out from this batch and is reported at 250- and 1{,}000-note scales. A
150-note subset of the 1{,}000-note ACT evaluation is additionally re-coded without
memory for a matched comparison. On MIMIC-IV, 25 LEARN notes construct memory and all
250 ACT notes are coded both with and without memory. We also report 500-note
base-pipeline benchmarks on each dataset without MistakeKDB.

\paragraph{Model and implementation.}
GPT-5 is used for all reported experiments. The implementation uses LangChain~\citep{LangChain},
JSON-structured outputs, CodeLookupDB, and at most three Coder--Judge rounds. All matched
comparisons hold the note set, model family, and pipeline structure fixed and vary the
presence of MistakeKDB.

\paragraph{Metrics.}
The primary metrics are \textbf{mean per-note (example-based) precision, recall, and
F1}: each metric is computed for a note's predicted and reference code sets and then
averaged across notes. This differs from the \emph{label-wise macro-F1} convention used
in much of the MIMIC extreme multi-label literature, which computes F1 independently for
each code label. Where indicated, we additionally report global micro metrics obtained
by pooling code decisions over the cohort. We retain the full available label space
rather than restricting evaluation to top-frequency labels.

\paragraph{Statistical comparisons.}
Our strongest controlled evidence for the memory effect comes from matched-note
comparisons. On 150 MIMIC-III ACT notes, we report paired-bootstrap 95\% confidence
intervals ($B{=}10{,}000$) and a Wilcoxon signed-rank test for per-note F1. On the
250-note MIMIC-IV ACT cohort, we report paired-bootstrap 95\% confidence intervals
($B{=}2{,}000$) and corresponding bootstrap $p$-values for within-note metric
differences. In the four-way MIMIC-III memory-format comparison, raw-example and
reflection-style conditions are reported as point estimates; the structured-versus-no-
memory contrast is the one accompanied by paired uncertainty and significance testing.
The 250- and 1{,}000-note memory-only MIMIC-III runs characterize the post-adaptation
operating point and scale stability; because they do not include no-memory predictions
for every note, they are not used by themselves to estimate a memory treatment effect.

\section{Results}
\label{sec:results}

\subsection{Matched-Note Effect of Mistake Memory}

\paragraph{MIMIC-III ($n{=}150$).}
The largest MIMIC-III matched comparison re-codes 150 ACT notes with and without
MistakeKDB (Table~\ref{tab:paired150}). Structured memory increases mean per-note CPT F1
from 11.5\% to 17.4\%, a +5.9 percentage-point difference (95\% CI
$[{+}3.3,{+}8.6]$; Wilcoxon $p{=}2.4\times10^{-5}$; 65 wins, 24 losses, 61 ties). ICD-9
moves from 41.1\% to 42.6\% (+1.5 points), with a confidence interval crossing zero
($[{-}0.3,{+}3.3]$, $p{=}0.11$). Thus, the strongest matched MIMIC-III evidence is a
relative improvement in procedural-code F1; it does not establish a significant ICD-9
F1 gain.

\begin{table}[t]
\centering
\small
\caption{Matched-note MistakeKDB ablation on 150 MIMIC-III ACT notes. Identical
  notes are coded with and without structured memory. Values are per-note F1; 95\% CIs
  use a paired bootstrap ($B{=}10{,}000$), and $p$-values use a Wilcoxon signed-rank
  test. W/L/T denotes per-note wins/losses/ties for memory.}
\label{tab:paired150}
\begin{tabular}{@{}l cc c c c c@{}}
\toprule
\textbf{Code} & \textbf{With} & \textbf{Without} & \textbf{$\Delta$F1}
  & \textbf{95\% CI} & \textbf{$p$} & \textbf{W/L/T} \\
\midrule
\textbf{CPT} & \textbf{17.4} & 11.5 & \textbf{+5.9} & $[{+}3.3, {+}8.6]$
  & $2.4{\times}10^{-5}$ & 65/24/61 \\
ICD-9 & 42.6 & 41.1 & +1.5 & $[{-}0.3, {+}3.3]$ & 0.11 & 76/66/8 \\
\bottomrule
\end{tabular}
\end{table}

\paragraph{MIMIC-IV ($n{=}250$).}
On the full matched MIMIC-IV ACT cohort, MistakeKDB increases ICD-10 precision by 4.2
points (95\% CI $[{+}2.0,{+}6.4]$, $p{=}0.003$) while decreasing recall by 3.1 points
(95\% CI $[{-}5.3,{-}0.9]$, $p{=}0.003$). The opposing changes leave F1 statistically
unchanged (+1.0 point, 95\% CI $[{-}1.0,{+}3.1]$, $p{=}0.30$). CPT/HCPCS F1 changes by
+1.8 points (95\% CI $[{-}2.1,{+}5.8]$, $p{=}0.36$), also not significant. The primary
analysis retains all 250 notes; excluding three notes with API-level extraction failures
leaves the significance conclusions unchanged. This result is important because it
shows what the memory mechanism reliably changes on this cohort: the ICD
precision--recall operating point, not overall F1.

\begin{table}[t]
\centering
\small
\caption{Matched-note MistakeKDB ablation on MIMIC-IV ($n{=}250$; ICD-10 + HCPCS).
  Identical ACT notes are coded with and without memory. Paired bootstrap 95\% CIs
  ($B{=}2{,}000$) quantify within-note differences.}
\label{tab:mimic4causal}
\begin{tabular}{@{}l c ccc ccc@{}}
\toprule
& & \multicolumn{3}{c}{\textbf{ICD-10}} & \multicolumn{3}{c}{\textbf{CPT/HCPCS}} \\
\cmidrule(lr){3-5} \cmidrule(lr){6-8}
\textbf{Condition} & \textbf{n} & \textbf{P} & \textbf{R} & \textbf{F1}
  & \textbf{P} & \textbf{R} & \textbf{F1} \\
\midrule
Without MistakeKDB & 250 & 36.1 & 48.0 & 39.8 & 12.6 & 16.9 & 13.6 \\
\textbf{With MistakeKDB} & \textbf{250} & \textbf{40.3} & \textbf{44.8} & \textbf{40.8}
  & \textbf{15.1} & \textbf{17.5} & \textbf{15.4} \\
\midrule
$\Delta$ (paired) & & +4.2$^{*}$ & $-$3.1$^{*}$ & +1.0 & +2.5 & +0.6 & +1.8 \\
\bottomrule
\end{tabular}
\begin{flushleft}
\footnotesize $^{*}p<0.01$ (paired bootstrap, $B{=}2{,}000$); unmarked deltas are not
significant at $p<0.05$.
\end{flushleft}
\end{table}

Taken together, the two matched cohorts argue against a uniform ``memory always improves
F1'' interpretation. Instead, the effect depends on code type and cohort: the largest
MIMIC-III paired effect is on CPT F1, while the clearest MIMIC-IV effect is an ICD
precision increase accompanied by lower recall. This heterogeneity motivates reporting
precision and recall separately and treating the role-aware memory as an adaptation
mechanism rather than a universal accuracy boost.

\subsection{Memory Representation on Matched MIMIC-III Notes}

To test whether the observed CPT gain arises from merely exposing the agents to prior-error
text or from how that experience is represented, we extend the same 150-note MIMIC-III
cohort with raw-example and reflection-style memory conditions. All four conditions use
the same underlying LEARN-phase error pool (1{,}028 mismatches), identical ACT notes,
the same model family, and the same inference pipeline; they differ in memory
representation (Table~\ref{tab:memfmt}). On CPT, structured MistakeKDB reaches 17.4\%
F1, compared with 11.5\% without memory, 12.1\% with raw examples, and 11.7\% with a
reflection summary. On ICD-9, all four formats lie within 1.5 points (41.1--42.6\%).
This pattern is consistent with structured representation being important for the
procedural-code improvement rather than the mere presence of prior-error text. However,
only the structured-versus-no-memory contrast is accompanied by paired confidence
intervals and significance testing (Table~\ref{tab:paired150}); raw and reflection
conditions are point estimates, so we do not claim statistically established superiority
over those two formats.

\begin{table}[t]
\centering
\small
\caption{Memory-format ablation on 150 matched MIMIC-III ACT notes. All four
  conditions code identical notes with an identical pipeline and the same underlying
  LEARN-phase error pool (1{,}028 mismatches); only the memory \emph{representation}
  differs. The no-memory and structured rows are the matched conditions tested in
  Table~\ref{tab:paired150}. Values are mean per-note precision, recall, and F1 (\%).}
\label{tab:memfmt}
\begin{tabular}{@{}l ccc ccc@{}}
\toprule
& \multicolumn{3}{c}{\textbf{ICD-9}} & \multicolumn{3}{c}{\textbf{CPT}} \\
\cmidrule(lr){2-4} \cmidrule(lr){5-7}
\textbf{Memory Format} & \textbf{P} & \textbf{R} & \textbf{F1}
  & \textbf{P} & \textbf{R} & \textbf{F1} \\
\midrule
No memory (baseline) & 41.3 & 44.0 & 41.1 & 11.0 & 14.1 & 11.5 \\
Raw examples & 43.1 & 45.0 & 42.5 & 13.0 & 14.0 & 12.1 \\
Reflection summary & 43.8 & 43.5 & 42.1 & 12.5 & 13.2 & 11.7 \\
\textbf{Structured MistakeKDB} & \textbf{43.3} & \textbf{44.8} & \textbf{42.6}
  & \textbf{20.8} & \textbf{19.7} & \textbf{17.4} \\
\midrule
$\Delta$ (Structured $-$ None) & +2.0 & +0.8 & +1.5
  & \textbf{+9.8} & \textbf{+5.6} & \textbf{+5.9} \\
$\Delta$ (Raw $-$ None)        & +1.8 & +1.0 & +1.4
  & +2.0 & $-$0.1 & +0.6 \\
\bottomrule
\end{tabular}
\end{table}

\subsection{Post-Adaptation Scale and Stability}

A 25-note MIMIC-III LEARN batch produces the MistakeKDB used throughout ACT evaluation.
On 250 held-out ACT notes, the memory-enabled system reaches 43.8\% mean per-note ICD-9
F1 and 19.1\% CPT F1; because there is no no-memory control for every note in this
cohort, this is an operating-point measurement rather than a treatment-effect estimate
(Appendix~\ref{app:secondary}). Extending the same memory to 1{,}000 held-out ACT notes
yields 43.7\% mean per-note ICD-9 F1 and 16.8\% CPT F1
(Table~\ref{tab:scale1000}). The near-identical ICD value at 250 and 1{,}000 notes
indicates scale stability under the same retrospective distribution, not a treatment
effect. The pipeline completed 999 of 1{,}000 coding calls successfully and required
1.91 debate rounds on average.

\begin{table}[t]
\centering
\small
\caption{Scale/stability evaluation on 1{,}000 held-out MIMIC-III ACT notes
  (ICD-9 + CPT), using the same 25-note MistakeKDB. Mean per-note and global micro
  precision, recall, and F1 are reported over the full available label space. The run
  completed successfully for 999/1{,}000 notes and converged in 1.91 debate rounds on average.}
\label{tab:scale1000}
\begin{tabular}{@{}l c ccc ccc@{}}
\toprule
& & \multicolumn{3}{c}{\textbf{ICD-9}} & \multicolumn{3}{c}{\textbf{CPT}} \\
\cmidrule(lr){3-5} \cmidrule(lr){6-8}
\textbf{ACT held-out (with MistakeKDB)} & \textbf{n} & \textbf{P} & \textbf{R} & \textbf{F1}
  & \textbf{P} & \textbf{R} & \textbf{F1} \\
\midrule
Mean per-note & 1{,}000 & 44.1 & 46.1 & \textbf{43.7} & 21.7 & 16.9 & \textbf{16.8} \\
Global micro & 1{,}000 & 44.3 & 43.4 & 43.8 & 20.6 & 18.3 & 19.4 \\
\bottomrule
\end{tabular}
\end{table}

Secondary base-pipeline, published-context, same-cohort component, and execution-time
analyses are reported in Appendix~\ref{app:secondary}. The first three are calibration
analyses rather than direct evidence for the memory effect; the latency analysis provides
complementary system-level efficiency measurements.

\subsection{Error Taxonomy and Mechanistic Interpretation}

Across 1{,}028 LEARN mismatches (full taxonomy in Appendix~\ref{app:secondary}, Table~\ref{tab:errors}), the MistakeAnalyzer assigns 68.2\% of LEARN mismatches to over-coding, under-coding,
wrong-code-type, or specificity categories---error types that map naturally to prompt-
level corrective actions. Over-coding lessons tighten Judge evidence requirements;
under-coding lessons expand Coder search; specificity errors motivate more explicit
acuity/laterality checks; and wrong-code-type errors flag billing-convention mismatches.
This operational mapping motivates role-aware memory, although the current taxonomy is
model-generated and has not been independently validated by professional coders.

\paragraph{Apparent documentation gaps (24.9\%).}
The MistakeAnalyzer labels 24.9\% of LEARN mismatches as documentation gaps, typically
when a reference administrative code appears unsupported by the discharge summary alone.
Examples include status/history codes or conditions recoverable from other parts of the
chart. We interpret these cases as an \emph{apparent single-document information ceiling},
not proof that the code is impossible to infer or that the reference label is clinically
perfect. They motivate chart-level retrieval and explicit separation between
note-supported coding and billing-record reconstruction.

\paragraph{Procedure-code failure modes.}
CPT/HCPCS errors frequently involve E\&M, critical-care, and facility-versus-professional
conventions. The significant relative CPT improvement in the 150-note matched MIMIC-III
analysis is encouraging, but absolute performance remains low. Procedure coding is
therefore both a setting where error memory can help and the clearest unresolved
limitation of the current single-note system.

\section{Discussion}
\label{sec:discussion}

\paragraph{What Learn-Then-Act changes.}
The results support cross-case error memory as an adaptation mechanism for a fixed
multi-agent workflow. On matched MIMIC-III notes, structured memory yields a significant
relative CPT F1 improvement, while the four-way representation comparison is consistent
with the organization of prior-error experience mattering for that gain. On MIMIC-IV,
the clearest effect is a precision increase paired with lower recall. These findings
therefore favor \emph{adaptive operating-point control} over a stronger claim of universal
accuracy improvement. The FN-to-Coder and FP-to-Judge routing is motivated by the
complementary recall/precision roles and is part of the evaluated system, but the routing
policy itself is not independently decomposed against symmetric or swapped-routing
controls.

\paragraph{Operational efficiency.}
In the matched MIMIC-III memory-format runs, structured MistakeKDB does not increase
observed wall-clock inference latency: it averages 239.8\,s/note and 1.83 debate rounds,
compared with 297.8\,s/note and 1.95 rounds without memory; raw-example and reflection
conditions are slower still (Appendix~\ref{app:latency}). Structured memory is also
associated with lower observed latency per round in these logs, so the latency difference
should not be attributed solely to fewer debate rounds. These measurements include
reasoning-model inference and API-gateway time on a single GPT-5 endpoint and are best
interpreted as system-level wall-clock measurements rather than isolated model-compute or
hardware-efficiency benchmarks.

\paragraph{Clinical interpretation and deployment scope.}
This study is retrospective and uses de-identified MIMIC data; \sys{} is not evaluated as
a deployed autonomous billing system or clinical decision-support tool. Administrative
codes are imperfect proxies for what is explicitly recoverable from a discharge note,
and lookup-table presence only verifies that an identifier exists in the loaded resource.
In a real coding workflow, mistake memories should be constructed from adjudicated audit
feedback, versioned with the applicable coding rules, and subject to human review before
reuse. The framework is intended to make such feedback reusable, not to remove coder
oversight.

\paragraph{Limitations.}
The controlled MIMIC-III analysis uses 150 matched notes, complemented by a larger
1{,}000-note ACT evaluation focused on memory stability rather than treatment effect.
CPT/HCPCS performance remains challenging, likely reflecting both model error and limited
chart and billing context. MistakeAnalyzer categories are intended as mechanistic labels
rather than adjudicated clinical annotations. The current study uses one model family and
does not yet examine model-version sensitivity, decoding stochasticity, stale or corrupted
memory, memory saturation, chart-level retrieval, or prospective workflow integration.
We report retrospective end-to-end inference latency, but do not evaluate deployment-scale
throughput, monetary cost, hardware-specific efficiency, or latency under a production
clinical infrastructure; the one-time LEARN-phase memory-construction cost is amortized
and excluded from the ACT-time measurements. In the four-way memory-format study, formal
paired uncertainty estimates are reported for structured versus no memory, while the
raw-example and reflection-style conditions are presented as point estimates. Additional
ablations of the FN-to-Coder / FP-to-Judge routing policy against symmetric or swapped
routing are left for future work.

\section{Conclusion}
\label{sec:conclusion}

Learn-Then-Act turns a small labeled feedback batch into role-specific error memory that can
be reused by a fixed clinical coding workflow without fine-tuning. Across matched
MIMIC-III and MIMIC-IV evaluations, memory does not produce a uniform F1 gain: structured
MistakeKDB yields a significant +5.9-point CPT F1 improvement on 150 MIMIC-III notes,
while MIMIC-IV ICD coding shifts toward higher precision at lower recall. On the same
150 MIMIC-III notes, raw-example and reflection memories remain near the no-memory CPT
baseline, a pattern consistent with structured representation being important for the
procedural-code gain. The same 25-note MIMIC-III memory also maintains a stable ICD-9
operating point when ACT evaluation is extended to 1{,}000 notes. These results support
error memory as a practical mechanism for adapting agent behavior from repeated audit
feedback while also exposing its limits: administrative labels can exceed single-note
evidence, procedure coding remains difficult, and clinical use would require adjudicated
feedback, richer context, and human oversight.

\clearpage

\clearpage
\appendix
\section{Secondary Evaluations and Published Context}
\label{app:secondary}

\subsection{MIMIC-III 250-Note Post-Adaptation Operating Point}

The 25-note LEARN batch and 250-note ACT cohort contain different notes. Table~\ref{tab:learn-act}
is therefore descriptive context only; the matched-note analyses in the main paper are
the basis for controlled claims about memory effects.

\begin{table}[t]
\centering
\small
\caption{Post-adaptation operating point on MIMIC-III (ICD-9 + CPT).
  The 25-note LEARN batch builds MistakeKDB and the 250-note ACT set is held out at the
  note level. The LEARN and ACT rows contain different notes and therefore must not be
  interpreted as a before/after effect estimate; memory effects are estimated only in
  matched-note ablations.}
\label{tab:learn-act}
\begin{tabular}{@{}l c ccc ccc@{}}
\toprule
& & \multicolumn{3}{c}{\textbf{ICD-9}} & \multicolumn{3}{c}{\textbf{CPT}} \\
\cmidrule(lr){3-5} \cmidrule(lr){6-8}
\textbf{Split} & \textbf{n} & \textbf{P} & \textbf{R} & \textbf{F1}
  & \textbf{P} & \textbf{R} & \textbf{F1} \\
\midrule
LEARN batch (no prior knowledge) & 25 & 42.8 & 45.3 & 42.6 & 17.4 & 14.4 & 15.4 \\
\textbf{ACT evaluation (with MistakeKDB)} & \textbf{250} & \textbf{45.8} & \textbf{46.9} & \textbf{43.8}
  & \textbf{21.5} & \textbf{21.1} & \textbf{19.1} \\
\bottomrule
\end{tabular}
\end{table}

\subsection{Base-Pipeline Benchmarks}

Without MistakeKDB, the base pipeline reaches 46.70\% mean per-note ICD-9 F1 on a
500-note MIMIC-III cohort and 40.98\% mean per-note ICD-10 F1 on a 500-note MIMIC-IV
cohort. Procedure-code performance is lower (25.79\% CPT on MIMIC-III and 12.93\%
CPT/HCPCS on MIMIC-IV), consistent with the stronger dependence of procedure labels on
billing context.

\begin{table}[t]
\centering
\small
\caption{Base-pipeline results without mistake memory on 500 discharge summaries per
  dataset. Metrics are mean per-note. MIMIC-III ICD-9 uses Judge review without direct
  ICD-9 lookup validation; MIMIC-IV ICD-10 has lookup-table coverage.}
\label{tab:main}
\begin{tabular}{@{}ll ccc ccc@{}}
\toprule
& & \multicolumn{3}{c}{\textbf{ICD}} & \multicolumn{3}{c}{\textbf{CPT/HCPCS}} \\
\cmidrule(lr){3-5} \cmidrule(lr){6-8}
\textbf{Dataset} & \textbf{Model} & \textbf{P} & \textbf{R} & \textbf{F1}
  & \textbf{P} & \textbf{R} & \textbf{F1} \\
\midrule
MIMIC-III (ICD-9, $n{=}500$) & GPT-5  & \textbf{52.21} & \textbf{48.27} & \textbf{46.70}
  & \textbf{29.23} & \textbf{25.67} & \textbf{25.79} \\
MIMIC-IV (ICD-10, $n{=}500$)  & GPT-5  & \textbf{37.67} & \textbf{48.86} & \textbf{40.98}
  & \textbf{12.43} & \textbf{14.16} & \textbf{12.93} \\
\bottomrule
\end{tabular}
\end{table}

\subsection{Published Coding Context}

Table~\ref{tab:sota} is intentionally not a direct ranking. Conventional MIMIC-III-FULL
papers report label-wise macro-F1 and global micro-F1, while our primary metric is mean
per-note F1 on sampled cohorts. MDACE additionally differs in note composition, ICD
version, and candidate space. The table is included only to situate the magnitude and
scope of recent supervised and agentic coding results.

\begin{table}[t]
\centering
\small
\caption{Published context for large-label-space clinical coding. The upper panel reports
  the standard \emph{label-wise} macro-F1 and global micro-F1 used on the conventional
  MIMIC-III-FULL benchmark; the lower panel reports recent agentic results on MDACE.
  These results are contextual only and are \textbf{not} a head-to-head ranking with our
  sampled MIMIC cohorts: our primary reported F1 is mean per-note (example-based), and
  MDACE uses ICD-10 with different note mixtures and candidate-code spaces. Values are
  taken from the cited papers.}
\label{tab:sota}
\resizebox{\textwidth}{!}{%
\begin{tabular}{@{}llccc@{}}
\toprule
\textbf{Method} & \textbf{Benchmark} & \textbf{Candidate space} & \textbf{Label-macro F1} & \textbf{Micro-F1} \\
\midrule
\multicolumn{5}{@{}l}{\emph{Standard MIMIC-III-FULL (ICD-9; conventional benchmark protocol)}} \\
CAML~\citep{Mullenbach2018}             & MIMIC-III-FULL & $\sim$8.9K & 8.8  & 53.9 \\
PLM-ICD~\citep{Huang2022PLMICD}         & MIMIC-III-FULL & $\sim$8.9K & 10.4 & 59.8 \\
ICDXML~\citep{Wang2024ICDXML}           & MIMIC-III-FULL & 8,922      & \textbf{18.3} & \textbf{62.5} \\
\midrule
\multicolumn{5}{@{}l}{\emph{Recent agentic coding on MDACE (ICD-10; different benchmark/protocol)}} \\
CLH-large~\citep{Huang2024CodeLikeHumans} & MDACE & $\sim$1K  & \textbf{28.0} & 43.0 \\
CLH-base (full ICD-10)~\citep{Huang2024CodeLikeHumans} & MDACE & $\sim$70K & 14.0 & 32.0 \\
MedDCR-GPT-5~\citep{Zheng2026MedDCR}     & MDACE & $\sim$1K  & --- & \textbf{51.0} \\
\bottomrule
\end{tabular}%
}
\end{table}

\paragraph{Lookup-table validity.}
Where direct lookup coverage is available, 99.86\% of predicted ICD-10 identifiers on
MIMIC-IV and 100\% of predicted CPT identifiers on both datasets are present in the
loaded reference tables. These percentages measure identifier presence, not clinical
correctness or billing appropriateness. We do not report an ICD-9 lookup-presence rate
because the loaded ICD database is ICD-10 based.

\subsection{Same-Cohort Component Ladder}

On the identical 250 MIMIC-IV ACT notes, no pairwise ICD or CPT F1 difference is
significant between the single-call Coder, Coder--Judge debate, and the full pipeline.
Debate shifts the operating point toward recall (lower precision, higher recall), and
MistakeKDB partially restores ICD precision. The ladder therefore supports a mechanism
of operating-point control rather than monotonic F1 improvement.

\begin{table}[t]
\centering
\small
\caption{Same-cohort component ladder on MIMIC-IV ($n{=}250$): single-call Coder,
  Coder--Judge debate without memory, and the full \sys{} pipeline. All rows use the
  identical held-out notes.}
\label{tab:ladder}
\begin{tabular}{@{}l ccc ccc@{}}
\toprule
& \multicolumn{3}{c}{\textbf{ICD-10}} & \multicolumn{3}{c}{\textbf{CPT/HCPCS}} \\
\cmidrule(lr){2-4} \cmidrule(lr){5-7}
\textbf{Rung} & \textbf{P} & \textbf{R} & \textbf{F1} & \textbf{P} & \textbf{R} & \textbf{F1} \\
\midrule
1. Zero-shot (no debate)       & 44.8 & 41.3 & 41.1 & 16.4 & 12.2 & 13.2 \\
2. Debate, no MistakeKDB       & 36.1 & 48.0 & 39.8 & 12.6 & 16.9 & 13.6 \\
\textbf{3. Full \sys{}}        & \textbf{40.3} & \textbf{44.8} & \textbf{40.8}
  & \textbf{15.1} & \textbf{17.5} & \textbf{15.4} \\
\bottomrule
\end{tabular}
\end{table}

\subsection{Execution Time and Overhead}
\label{app:latency}

We report wall-clock cost of the Coder--Judge pipeline. Per-note latency and debate
rounds are logged for every note; the per-step breakdown in Table~\ref{tab:steptime} is
reconstructed from timestamped log events and therefore reflects end-to-end call latency
(reasoning-model inference plus API-gateway time), not isolated model compute. All runs
use the same GPT-5 reasoning endpoint. The one-time LEARN-phase MistakeKDB construction
(25 notes analyzed by the MistakeAnalyzer) is amortized across all ACT notes and is not
included in these ACT-time figures.

Two observations follow. First, memory \emph{format} is associated with meaningful
wall-clock differences: structured MistakeKDB is the fastest matched configuration
(239.8\,s/note) and is associated with both fewer debate rounds (1.83) and lower observed
latency per round than the no-memory, raw-example, and reflection conditions
(Table~\ref{tab:latency}). Because both round count and per-round latency vary across
conditions, these logs do not isolate the cause of the latency difference. Second, the
initial recall-oriented Coder extraction is the most expensive single step
($\sim$102\,s), revision passes are $\sim$40\% cheaper, and the Judge is comparable to an
average Coder call (Table~\ref{tab:steptime}).

\begin{table}[t]
\centering
\small
\caption{End-to-end latency and debate cost per note. Latency and rounds are logged for
  every note; latency/round is the derived mean, where each round issues two LLM calls
  (one Coder, one Judge). Matched rows use the 150-note MIMIC-III cohort of
  Table~\ref{tab:memfmt}.}
\label{tab:latency}
\begin{tabular}{@{}l c cc c@{}}
\toprule
\textbf{Configuration} & \textbf{n} & \textbf{Latency/note (s)} & \textbf{Rounds}
  & \textbf{Latency/round (s)} \\
\midrule
Structured (scale run)      & 1{,}000 & 265.4 & 1.91 & 138.9 \\
\textbf{Structured} (matched) & 150   & \textbf{239.8} & \textbf{1.83} & 131.3 \\
No memory (matched)         & 150     & 297.8 & 1.95 & 153.0 \\
Raw (matched)               & 150     & 314.3 & 1.99 & 158.2 \\
Reflection (matched)        & 150     & 329.9 & 1.96 & 168.3 \\
\bottomrule
\end{tabular}
\end{table}

\begin{table}[t]
\centering
\small
\caption{Per-LLM-call step durations, reconstructed from timestamped log events over the
  matched memory-format runs (raw + reflection arms; 633 Coder calls and 604 Judge
  calls). Durations are end-to-end (model inference plus gateway latency).}
\label{tab:steptime}
\begin{tabular}{@{}l r ccc@{}}
\toprule
\textbf{Step} & \textbf{Calls} & \textbf{Mean (s)} & \textbf{Median (s)} & \textbf{p90 (s)} \\
\midrule
Coder --- extract (round 1)   & 329 & 102.1 & 96.6 & 141.0 \\
Coder --- revise (round $\geq$2) & 304 & 61.6 & 58.3 & 83.8 \\
Coder --- all calls           & 633 & 82.6 & 78.4 & 123.6 \\
Judge --- review (all)        & 604 & 81.9 & 76.1 & 110.9 \\
\bottomrule
\end{tabular}
\end{table}

\subsection{Full MistakeAnalyzer Taxonomy}

The taxonomy below underlies the mechanistic summary in the main paper. It is generated
by the MistakeAnalyzer and has not been independently adjudicated by professional coders.

\begin{table}[t]
\centering
\small
\caption{MistakeAnalyzer taxonomy over 1,028 prediction/reference-label mismatches
  from the MIMIC-III LEARN phase. Categories marked with $\star$ are directly targeted
  by the current memory design; the taxonomy has not been independently coder-adjudicated.}
\label{tab:errors}
\begin{tabular}{@{}lrrrr@{}}
\toprule
\textbf{Category} & \textbf{FP} & \textbf{FN} & \textbf{Total (\%)} & \textbf{Learnable?} \\
\midrule
Documentation gap  &  13 & 243 & 256 (24.9) & \textsf{---} \\
Over-coding        & 227 &   0 & 227 (22.1) & $\star$ \\
Under-coding       &   0 & 211 & 211 (20.5) & $\star$ \\
Wrong code type    & 110 &  23 & 133 (12.9) & $\star$ \\
Specificity        &  64 &  67 & 131 (12.7) & $\star$ \\
Other / invalid    &  63 &   7 &  70 \phantom{0}(6.8) & partial \\
\midrule
\textbf{Total}     & \textbf{477} & \textbf{551} & \textbf{1,028} & \\
\bottomrule
\end{tabular}
\end{table}


\begin{thebibliography}{99}

\bibitem[Boukhers et~al.(2024)]{boukhers2024large}
Zeyd Boukhers, Ameer Ali Khan, Qusai Ramadan, and Cong Yang.
\newblock Large language model in medical informatics: Direct classification and enhanced text representations for automatic ICD coding.
\newblock In \emph{2024 IEEE International Conference on Bioinformatics and Biomedicine (BIBM)}, pages 3066--3069. IEEE, 2024.
\newblock \href{https://doi.org/10.1109/BIBM62325.2024.10822419}{doi:10.1109/BIBM62325.2024.10822419}.

\bibitem[Boyle et~al.(2023)]{Boyle2023GPTCoding}
Joseph S. Boyle, Antanas Kascenas, Pat Lok, Maria Liakata, and Alison Q. O'Neil.
\newblock Automated clinical coding using off-the-shelf large language models.
\newblock In \emph{NeurIPS 2023 Workshop on Deep Generative Models for Health}, 2023.
\newblock \href{https://arxiv.org/abs/2310.06552}{arXiv:2310.06552}.

\bibitem[Du et~al.(2023)]{Du2023Debate}
Yilun Du, Shuang Li, Antonio Torralba, Joshua B. Tenenbaum, and Igor Mordatch.
\newblock Improving factuality and reasoning in language models through multiagent debate.
\newblock \emph{arXiv preprint}, 2023.
\newblock \href{https://arxiv.org/abs/2305.14325}{arXiv:2305.14325}.

\bibitem[Gomes et~al.(2024)]{gomes2024accurate}
Goncalo Gomes, Isabel Coutinho, and Bruno Martins.
\newblock Accurate and well-calibrated ICD code assignment through attention over diverse label embeddings.
\newblock In \emph{Proceedings of the 18th Conference of the European Chapter of the Association for Computational Linguistics}, pages 2302--2315, 2024.
\newblock \href{https://doi.org/10.18653/v1/2024.eacl-long.141}{doi:10.18653/v1/2024.eacl-long.141}.

\bibitem[Huang et~al.(2022)]{Huang2022PLMICD}
Chao-Wei Huang, Shang-Chi Tsai, and Yun-Nung Chen.
\newblock PLM-ICD: Automatic ICD coding with pretrained language models.
\newblock In \emph{Proceedings of the 4th Clinical Natural Language Processing Workshop}, pages 10--20. Association for Computational Linguistics, 2022.
\newblock \href{https://doi.org/10.18653/v1/2022.clinicalnlp-1.2}{doi:10.18653/v1/2022.clinicalnlp-1.2}.

\bibitem[Wang et~al.(2024)]{Wang2024ICDXML}
Zeqiang Wang, Yuqi Wang, Haiyang Zhang, Wei Wang, Jun Qi, Jianjun Chen, Nishanth Sastry, Jon Johnson, and Suparna De.
\newblock ICDXML: Enhancing ICD coding with probabilistic label trees and dynamic semantic representations.
\newblock \emph{Scientific Reports}, 14:18319, 2024.
\newblock \href{https://doi.org/10.1038/s41598-024-69214-9}{doi:10.1038/s41598-024-69214-9}.

\bibitem[Ji et~al.(2023)]{Ji2023Hallucination}
Ziwei Ji, Nayeon Lee, Rita Frieske, Tiezheng Yu, Dan Su, Yan Xu, Etsuko Ishii, Ye Jin Bang, Andrea Madotto, and Pascale Fung.
\newblock Survey of hallucination in natural language generation.
\newblock \emph{ACM Computing Surveys}, 55(12):1--38, 2023.
\newblock \href{https://doi.org/10.1145/3571730}{doi:10.1145/3571730}.

\bibitem[Johnson et~al.(2016)]{Johnson2016MIMICIII}
Alistair E. W. Johnson, Tom J. Pollard, Lu Shen, Li-wei H. Lehman, Mengling Feng, Mohammad Ghassemi, Benjamin Moody, Peter Szolovits, Leo Anthony Celi, and Roger G. Mark.
\newblock MIMIC-III, a freely accessible critical care database.
\newblock \emph{Scientific Data}, 3:160035, 2016.
\newblock \href{https://doi.org/10.1038/sdata.2016.35}{doi:10.1038/sdata.2016.35}.

\bibitem[Johnson et~al.(2023)]{Johnson2023MIMICIV}
Alistair E. W. Johnson, Lucas Bulgarelli, Lu Shen, Alvin Gayles, Ayad Shammout, Steven Horng, Tom J. Pollard, Sicheng Hao, Benjamin Moody, Brian Gow, Li-wei H. Lehman, Leo Anthony Celi, and Roger G. Mark.
\newblock MIMIC-IV, a freely accessible electronic health record dataset.
\newblock \emph{Scientific Data}, 10:1, 2023.
\newblock \href{https://doi.org/10.1038/s41597-022-01899-x}{doi:10.1038/s41597-022-01899-x}.

\bibitem[Chase(2022)]{LangChain}
Harrison Chase.
\newblock LangChain.
\newblock Software framework, 2022.
\newblock \url{https://github.com/langchain-ai/langchain}.

\bibitem[Le et~al.(2025)]{Ren2025ACEICD}
Tuan-Dung Le, Shohreh Haddadan, and Thanh Q. Thieu.
\newblock ACE-ICD: Acronym expansion as data augmentation for automated ICD coding.
\newblock In \emph{Findings of the Association for Computational Linguistics: IJCNLP-AACL 2025}, pages 1650--1662, 2025.

\bibitem[Li and Yu(2020)]{Li2020MultiResCNN}
Fei Li and Hong Yu.
\newblock ICD coding from clinical text using multi-filter residual convolutional neural network.
\newblock In \emph{Proceedings of the AAAI Conference on Artificial Intelligence}, volume 34, pages 8180--8187, 2020.
\newblock \href{https://doi.org/10.1609/aaai.v34i05.6331}{doi:10.1609/aaai.v34i05.6331}.

\bibitem[Li et~al.(2024)]{Huang2024LLMCoding}
Rumeng Li, Xun Wang, and Hong Yu.
\newblock Exploring LLM multi-agents for ICD coding.
\newblock \emph{arXiv preprint}, 2024.
\newblock \href{https://arxiv.org/abs/2406.15363}{arXiv:2406.15363}.

\bibitem[Liang et~al.(2023)]{Liang2023Divergent}
Tian Liang, Zhiwei He, Wenxiang Jiao, Xing Wang, Yan Wang, Rui Wang, Yujiu Yang, Zhaopeng Tu, and Shuming Shi.
\newblock Encouraging divergent thinking in large language models through multi-agent debate.
\newblock \emph{arXiv preprint}, 2023.
\newblock \href{https://arxiv.org/abs/2305.19118}{arXiv:2305.19118}.

\bibitem[Madaan et~al.(2023)]{Madaan2023SelfRefine}
Aman Madaan, Niket Tandon, Prakhar Gupta, Skyler Hallinan, Luyu Gao, Sarah Wiegreffe, Uri Alon, Nouha Dziri, Shrimai Prabhumoye, Yiming Yang, Shashank Gupta, Bodhisattwa Prasad Majumder, Katherine Hermann, Sean Welleck, Amir Yazdanbakhsh, and Peter Clark.
\newblock Self-refine: Iterative refinement with self-feedback.
\newblock \emph{arXiv preprint}, 2023.
\newblock \href{https://arxiv.org/abs/2303.17651}{arXiv:2303.17651}.

\bibitem[Motzfeldt et~al.(2025)]{Huang2024CodeLikeHumans}
Andreas Geert Motzfeldt, Joakim Edin, Casper L. Christensen, Christian Hardmeier, Lars Maaloe, and Anna Rogers.
\newblock Code like humans: A multi-agent solution for medical coding.
\newblock In \emph{Findings of the Association for Computational Linguistics: EMNLP 2025}, pages 22612--22627, 2025.
\newblock \href{https://doi.org/10.18653/v1/2025.findings-emnlp.1231}{doi:10.18653/v1/2025.findings-emnlp.1231}.

\bibitem[Mullenbach et~al.(2018)]{Mullenbach2018}
James Mullenbach, Sarah Wiegreffe, Jon Duke, Jimeng Sun, and Jacob Eisenstein.
\newblock Explainable prediction of medical codes from clinical text.
\newblock In \emph{Proceedings of the 2018 Conference of the North American Chapter of the Association for Computational Linguistics: Human Language Technologies}, pages 1101--1111, 2018.
\newblock \href{https://doi.org/10.18653/v1/N18-1100}{doi:10.18653/v1/N18-1100}.

\bibitem[O'Malley et~al.(2005)]{OMalley2005}
Kimberly J. O'Malley, Kevin F. Cook, Matthew D. Price, Kimberly R. Wildes, John F. Hurdle, and Carol M. Ashton.
\newblock Measuring diagnoses: ICD code accuracy.
\newblock \emph{Health Services Research}, 40(5 Pt 2):1620--1639, 2005.
\newblock \href{https://doi.org/10.1111/j.1475-6773.2005.00444.x}{doi:10.1111/j.1475-6773.2005.00444.x}.

\bibitem[Pal et~al.(2023)]{Pal2023MedHALT}
Ankit Pal, Logesh Kumar Umapathi, and Malaikannan Sankarasubbu.
\newblock Med-HALT: Medical Domain Hallucination Test for Large Language Models.
\newblock In \emph{Proceedings of the 27th Conference on Computational Natural Language Learning (CoNLL)}, pages 314--334, 2023.
\newblock \href{https://doi.org/10.18653/v1/2023.conll-1.21}{doi:10.18653/v1/2023.conll-1.21}.

\bibitem[Shinn et~al.(2023)]{Shinn2023Reflexion}
Noah Shinn, Federico Cassano, Edward Berman, Ashwin Gopinath, Karthik Narasimhan, and Shunyu Yao.
\newblock Reflexion: Language agents with verbal reinforcement learning.
\newblock \emph{arXiv preprint}, 2023.
\newblock \href{https://arxiv.org/abs/2303.11366}{arXiv:2303.11366}.

\bibitem[Tang et~al.(2024)]{Tang2024MedAgents}
Xiangru Tang, Anni Zou, Zhuosheng Zhang, Ziming Li, Yilun Zhao, Xingyao Zhang, Arman Cohan, and Mark Gerstein.
\newblock MedAgents: Large Language Models as Collaborators for Zero-shot Medical Reasoning.
\newblock In \emph{Findings of the Association for Computational Linguistics: ACL 2024}, pages 599--621, 2024.
\newblock \href{https://doi.org/10.18653/v1/2024.findings-acl.33}{doi:10.18653/v1/2024.findings-acl.33}.

\bibitem[Vu et~al.(2020)]{Vu2020LAAT}
Thanh Vu, Dat Quoc Nguyen, and Anthony Nguyen.
\newblock A label attention model for ICD coding from clinical text.
\newblock In \emph{Proceedings of the Twenty-Ninth International Joint Conference on Artificial Intelligence}, pages 3335--3341, 2020.
\newblock \href{https://doi.org/10.24963/ijcai.2020/461}{doi:10.24963/ijcai.2020/461}.

\bibitem[Wu et~al.(2025)]{RoSimTail2025}
Yuhao Wu, Yifan Wu, Wei Fan, and Min Li.
\newblock Enhancing ICD classification with semantic embedding rectification and long-tail refinement.
\newblock \emph{Knowledge-Based Systems}, 330:114530, 2025.
\newblock \href{https://doi.org/10.1016/j.knosys.2025.114530}{doi:10.1016/j.knosys.2025.114530}.

\bibitem[xerk-dot(2024)]{Xerk2024MedCodingAI}
xerk-dot.
\newblock Medical coding AI: A comprehensive benchmarking platform for CPT, ICD-10, and HCPCS coding questions.
\newblock Software repository, 2024.
\newblock \url{https://github.com/xerk-dot/medical-coding-ai}.

\bibitem[Yang et~al.(2023)]{yang2023surpassing}
Zhichao Yang, Sanjit Singh Batra, Joel Stremmel, and Eran Halperin.
\newblock Surpassing GPT-4 medical coding with a two-stage approach.
\newblock \emph{arXiv preprint}, 2023.
\newblock \href{https://arxiv.org/abs/2311.13735}{arXiv:2311.13735}.

\bibitem[Yang et~al.(2022)]{Yang2023KEPT}
Zhichao Yang, Shufan Wang, Bhanu Pratap Singh Rawat, Avijit Mitra, and Hong Yu.
\newblock Knowledge injected prompt based fine-tuning for multi-label few-shot ICD coding.
\newblock In \emph{Findings of the Association for Computational Linguistics: EMNLP 2022}, 2022.
\newblock \href{https://arxiv.org/abs/2210.03304}{arXiv:2210.03304}.

\bibitem[Yuan et~al.(2022)]{Yuan2022MSMN}
Zheng Yuan, Chuanqi Tan, and Songfang Huang.
\newblock Code synonyms do matter: Multiple synonyms matching network for automatic ICD coding.
\newblock In \emph{Proceedings of the 60th Annual Meeting of the Association for Computational Linguistics}, pages 808--814, 2022.
\newblock \href{https://doi.org/10.18653/v1/2022.acl-short.91}{doi:10.18653/v1/2022.acl-short.91}.

\bibitem[Zheng et~al.(2026)]{Zheng2026MedDCR}
Jiyang Zheng, Islam Nassar, Thanh Vu, Xu Zhong, Yang Lin, Tongliang Liu, Long Duong, and Yuan-Fang Li.
\newblock MedDCR: Learning to design agentic workflows for medical coding.
\newblock In \emph{Findings of the Association for Computational Linguistics: ACL 2026}, pages 12878--12893, 2026.
\newblock \href{https://doi.org/10.18653/v1/2026.findings-acl.627}{doi:10.18653/v1/2026.findings-acl.627}.

\bibitem[Zhao et~al.(2023)]{Zhao2024ExpeL}
Andrew Zhao, Daniel Huang, Quentin Xu, Matthieu Lin, Yong-Jin Liu, and Gao Huang.
\newblock ExpeL: LLM agents are experiential learners.
\newblock \emph{arXiv preprint}, 2023.
\newblock \href{https://arxiv.org/abs/2308.10144}{arXiv:2308.10144}.

\end{thebibliography}
\end{document}